# Data preprocessing methods for robust Fourier ptychographic microscopy

Yan Zhang[1,2,3], An Pan[1,2,3], Ming Lei[1], and Baoli Yao[1,*]

[1] State Key Laboratory of Transient Optics and Photonics, Xi'an Institute of Optics and Precision Mechanics, Chinese Academy of Sciences, Xi'an 710119, China
[2] University of Chinese Academy of Sciences, Beijing 100049, China
[3] These authors contribute equally to this paper
*Corresponding author: yaobl@opt.ac.cn

Fourier ptychographic microscopy (FPM) is a recently proposed computational imaging technique with both high resolution and wide field-of-view. In current FP experimental setup, the dark-field images with high-angle illuminations are easily submerged by stray light and background noise due to the low signal-to-noise ratio, thus significantly degrading the reconstruction quality and also imposing a major restriction on the synthetic numerical aperture (NA) of the FP approach. To this end, an overall and systematic data preprocessing scheme for noise removal from FP's raw dataset is provided, which involves sampling analysis as well as underexposed/overexposed treatments, then followed by the elimination of unknown stray light and suppression of inevitable background noise, especially Gaussian noise and CCD dark current in our experiments. The reported non-parametric scheme facilitates great enhancements of the FP's performance, which has been demonstrated experimentally that the benefits of noise removal by these methods far outweigh its defects of concomitant signal loss. In addition, it could be flexibly cooperated with the existing state-of-the-art algorithms, producing a stronger robustness of the FP approach in various applications.

## 1. INTRODUCTION

Fourier ptychography microscopy (FPM) [1-3] is a recently developed computational imaging technique that applies angular diversity to recover a large field of view (FOV), high-resolution (HR) complex image. Sharing its roots with synthetic aperture technique, FPM iteratively synthesizes a number of variably-illuminated low-resolution (LR) images to expand the detected field's Fourier passband, which makes the final achievable resolution determined by the sum of objective lens and illumination NAs [4]. Due to its flexible setup, perfect performance and rich redundancy of recorded data, FPM has found wide applications in the study of 3D imaging [5], high-resolution fluorescence imaging [6, 7], multiplexing imaging [8, 9] and high-speed in vitro imaging [10].

In current FP's setup, due to the low signal-to-noise ratio (SNR), the dark-field (DF) images with high-angle illuminations could easily be submerged by unknown stray light and inevitable background noise, including CCD dark current, thermal or readout noise, hot pixels, cosmic rays, etc. They severely break the consistencies between the captured intensity images, causing the iterative reconstruction algorithms never truly converge to a unique solution [11]. In recent years, a lot of the state-of-the-art algorithms [12-16] have been proposed to suppress such noises in some specific cases. However, on the one hand, these targeted methods generally reside on expensive algorithm complexity, making them less appealing from a computational point of view [14]; On the other hand, they usually require a priori knowledge about noise statistics, such as constant background noise (uniform measurement bias) [17], Gaussian or Poisson distributed noise [18, 19], and fluctuating background noise that varies with pixels but keeps constant distribution throughout the process of data acquisitions [20]. While in practice, some complex noise cannot be accurately expressed by any of the models mentioned above, thus imposing a great restriction on their applications.

For computational imaging technique, data preprocessing is the most fundamental and indispensable step before image reconstruction, which also provides a traditional way for noise suppression. However, this step is somewhat ambiguous because some parts of the meaningful signals are inevitably wiped out along with the noise. For this reason, an integral data preprocessing scheme for FPM has been rarely reported, which is often one-sidedly explained or even unmentioned in recent publications. In this work, we summarized and also newly proposed a series of methods for noise removal from the FP's raw dataset. Firstly, a sampling analysis and the sparsely-sampled scheme [21] are applied to address under-sampling as well as underexposed/overexposed issues. Then a binary mask is introduced in updating process to eliminate the negative effects of stray light, and some hot pixels can also be replaced by the average of their adjacent pixels. Next, a weighted subtraction of CCD dark image is employed to uniformize the noise distribution, which contributes a lot to Gaussian noise. Finally, a thresholding method is implemented to suppress the residual background noise that mainly comes from CCD dark current. All these steps form an overall and systematic data preprocessing scheme, which doesn't require any parameter settings and priori knowledge about noise statistical properties. In addition, it can be further cooperated with different kinds of state-of-the-art algorithms, facilitating a better performance in final reconstructions. The experimental results indicated that the benefits of noise removal by these methods far outweigh its defects of accompanying signal loss, because part of the lost signals can be compensated by the improved consistencies between the captured raw images.

## 2. DATA PREPROCESSING METHODS

In this section, an overall and systematic data preprocessing scheme is provided, which involves sampling analysis as well as underexposed/overexposed treatments, then followed by the stray light elimination and background noise suppression, especially Gaussian noise and CCD dark current. Note that the intense noise introduced by human factors would affect solution convergence. Therefore, elaborate experiments are still needed to ensure a successful image reconstruction.

The proposed data preprocessing scheme is starting with a discussion on the undersampling issue. According to sampling theorem, a pixel size larger than $\lambda/(2NA_{obj})$, where $\lambda$ is the excitation wavelength and $NA_{obj}$ is the numerical aperture of the employed objective, may lead to a pixel aliasing problem in the Fourier domain, greatly degrading the FP's reconstruction quality. To address this problem, a sub-sampled mask [21] is employed to divide each pixel into several sub-pixels, making the effective pixel size reduced by half of the original one. Then, the dataset is processed with the sparsely-sampled scheme [21] to eliminate the underexposed/overexposed pixels, which are attributed to the limited bit-depth of image sensors. This scheme produces a comparable recovery quality to that with a high-dynamic-range (HDR) combination process, thus avoiding the multi-exposure acquisitions for synthesizing a HDR image.

After the above treatments, the data preprocessing scheme is then followed by stray light elimination and noise suppression, which will be detailed in the following sections A, B and C.

### A.  Elimination of Stray Light

In refractive optics, stray light mostly comes from the multi-reflections between lenses, scattering of internal baffles, or imperfection of optical elements [22]. It could be partly eliminated by employing lens hood or black coating, but the residual stray light inevitably remains in the experimental system, which would increase the local brightness of dark-field (DF) images as shown in Fig. 1 group (b), thus severally breaking the consistencies between the captured images. It's worth noting that the stray light is distributed randomly, therefore, some affected DF images are obvious while others are less visible in practice. The proposed method enables the ability to automatically pick the affected images from experimental raw dataset, and then eliminate their negative effects in reconstruction process.

Figure 1 illustrates the principle of the proposed method. Fig. 1(a) shows the overlap of sub-apertures due to the angle-varied illuminations in the Fourier domain. (a1) and (a2) demonstrate a bright-field (BF) image and a dark-filed (DF) image respectively, with (a1) within while (a2) beyond the numerical aperture of the employed objective. The corresponding histograms also indicate a great difference on their intensity distributions, where the average intensity of a BF image is several orders of magnitude higher than that of a DF image. Actually, the normalized intensity of a DF image is no more than 0.1 in most cases, unless it is affected by the stray light, which would increase its local brightness. On the basis of this analysis, an automatic identification of the affected DF images could be achieved by introducing an intensity threshold $\eta$, and the cases where the intensity of simulated image is lower than $\eta$ but the real intensity measurement is greater than $\eta$ will be considered as a corrupt image that affected by the stray light shown in Fig.1(b) as examples.

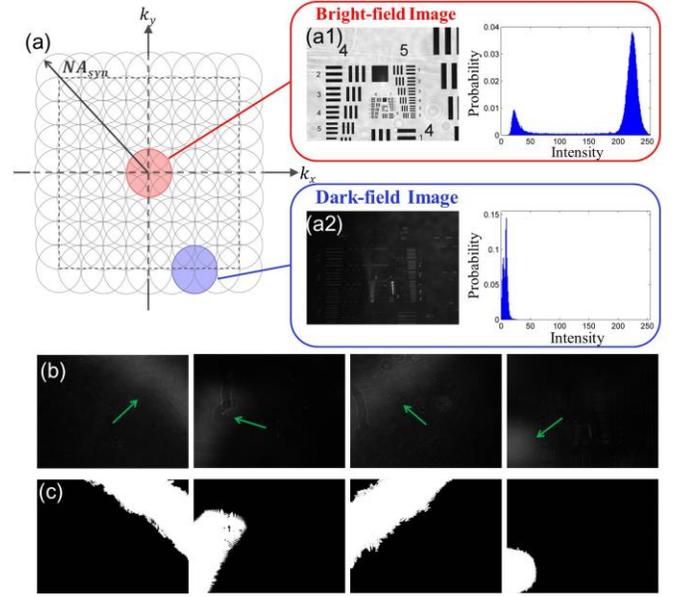

**Fig. 1**  Principle of the stray light elimination. (a) Overlap of sub-apertures in spectrum space; (a1) A bright-field image and (a2) A dark-filed image, and their corresponding histograms. Both are captured with a 4×, 0.1 NA objective lens and an 8-bit CCD camera. (b) A series of corrupt images that affected by the stray light. (c) The generated binary masks correspond to group (b).

Next, we binarize those affected images by employing Otsu's method [24], which minimizes the intraclass variance of the black and white pixels, and maximally extracts the affected areas from background signals. Group (c) presents the generated binary masks that correspond to group (b). Finally, we apply these binary masks in the amplitude updating step of EPRY-FPM algorithm [23], where the modulus of simulated images are replaced by the square-root of intensity measurements. During this updating, the regions affected by stray light will be retained unchanged while other regions will be updated by the captured raw images. As a consequence, the amplitude updating step would be modified as follows:

$$\phi_i^{upd}(x,y) = \sqrt{I_i^{upd}(x,y)} \frac{\phi_i^e(x,y)}{\left|\phi_i^e(x,y)\right|} \quad (1)$$

In Eq. (1), subscript $i$ corresponds to the $i^{th}$ illumination angle, and $(x,y)$ denotes the 2D spatial coordinates in the image plane. $\phi_i^e(x,y)$ presents the complex amplitude of simulated emission light on the CCD plane, and $I_i^{upd}(x,y)$ is the updated intensity measurements, which can be expressed as follows:

$$I_i^{upd}(x,y) = \begin{cases} I_i^c(x,y), & |\phi_i^e(x,y)|^2 \leq \eta \ \& \ I_i^c(x,y) \leq \eta \\ m_i(x,y) \cdot |\phi_i^e(x,y)|^2 + [1-m_i(x,y)] \cdot I_i^c(x,y), & else \end{cases} \quad (2)$$

Where $I_i^c(x,y)$ presents the normalized intensity of the $i^{th}$ captured image, while $m_i(x,y)$ denotes the corresponding binary mask such as Fig. 1 Group (c), and $\eta$ is the normalized intensity threshold. Eq. (2) enables the ability to eliminate the negative effects of stray light in reconstruction process. In view of the great difference of intensity distributions between the BF and DF images, there is a wide range of options for threshold selection. For non-parametric design, the normalized intensity of 0.1 is usually taken as the threshold with empirics.

Depending on the percentage of invalid pixels, one may need to increase the number of plane wave illuminations to ensure the solution convergence. As a reference point [21], the percentage of the affected pixels is typically no more than 15% in the FP's experiment.

### B. Uniformity Method

The uneven distribution of background noise can be observed in a CCD dark image, which is a mixture of Gaussian or Poisson noise, CCD dark current, thermal or readout noise, etc. They also break the consistencies between the raw images and degrade the FP's recovery quality. To further improve the consistencies, a weighted subtraction of CCD dark image is employed to uniformize the noise distribution, which is particularly helpful to Gaussian noise.

Figure 2 shows one of the false-color raw images of a USAF target captured with a 4×, 0.1NA objective lens and an 8-bit CCD camera with a pixel size of 3.75$\mu m$. As demonstrated by red boxes in Fig. 2(a), two rectangular sub-regions, R1 (300×300 pixels) and R2 (200×200 pixels), which contain a relatively high level of background noise and low target signals compared with other regions, are selected to calculate a weighting factor for the weighted subtraction. Meanwhile, the same areas in the CCD dark image are also selected as shown in Fig. 2 (b). Here (a1), (a2) and (b1), (b2) denote the corresponding zoom-in images of two sub-regions. Next, the weighting factor for the ith raw image is calculated as follows:

$$\alpha_i = \frac{\left\langle \sum_{x,y \in R_1, R_2} I_{M,i}(x,y) I_D(x,y) \right\rangle}{\left\langle \sum_{x,y \in R_1, R_2} I_D(x,y)^2 \right\rangle} \quad (3)$$

Where $I_{M,i}(x,y)$ and $I_D(x,y)$ present the intensity of the $i^{th}$ measured image and CCD dark image respectively, and $(x,y)$ denotes the 2D spatial coordinates in the image plane. The operator $\langle \cdot \rangle$ denotes the mean value of two sub-regions. Equation (3) minimizes the difference between the measured images and CCD dark image by employing the least squares method. Then a weighted subtraction is carried out to update the intensity of the $i^{th}$ measured image:

$$I_i^u(x,y) = I_{M,i}(x,y) - \alpha_i I_D(x,y) \quad (4)$$

Equation (4) produces the improved consistencies between the captured raw images, which varies in the magnitude of background noise due to different exposure time and beam instability during the data acquisitions. It's worth mentioning that the appropriately selected regions, such as their numbers, size and positions, could partly account for the uneven distribution of background noise within one captured image. To better reflect the statistical properties of background noise, the mean value of multiple regions could be employed to calculate the weighting factor. In general, two chosen regions are often sufficient in the real situations.

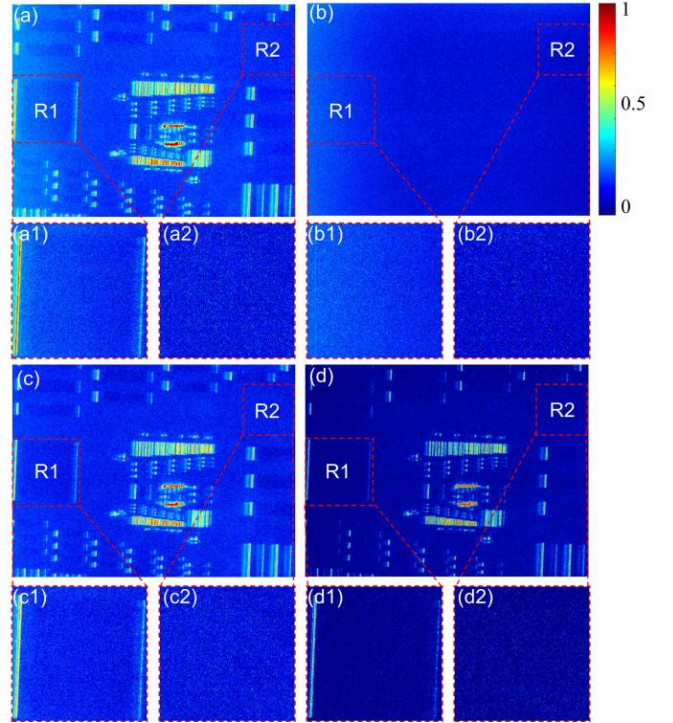

**Fig. 2** Uniformity method for background noise removal. (a) One of the false-color raw images of a USAF target, captured with a 4×, 0.1NA objective lens and an 8-bit CCD camera. (b) CCD dark image averaged by 20 exposures. (c) Target image obtained by subtracting (b) from (a) directly. (d) Target image obtained by employing the uniformity method on (a) and (b). (a1)~(d1) and (a2)~(d2) The corresponding zoom-in images of two sub-regions R1 (300×300 pixels) and R2 (200×200 pixels).

In most cases, a clean and uniform target pattern could be obtained after the uniformity method, which is indicated in Fig.2 (d). As a comparison, a direct subtraction of CCD dark image (equivalent to $\alpha_i=1$) is also demonstrated in Fig. 2(c). Through a comparison of the zoom-in images as shown in (c1), (c2) and (d1), (d2), it can be observed that Fig. 2(d) has a much cleaner background with a lower and more uniform noise distribution than Fig. 2(c), which validates the effectiveness of the reported method. However, in some cases, a handful of residual noise may still exist in the experimental dataset after this step. Considering this situation, a thresholding method will be introduced in the following section.

### C. Thresholding Method

After implemented by step B, the dataset is then processed with the thresholding method to suppress residual background noise,

such as the constantly distributed CCD dark current, which is always related to the thermal motion of carriers. In this case, an appropriate threshold value $I_{th}$ is introduced for image segmentation, the images whose intensity is lower than $I_{th}$ will be set to zero. It is worth mentioning that a reasonable threshold is crucial to this method, as a low value will leave more residual noise while a high value will erase more meaningful signals. Here, as a reference, an empirical threshold value [25] can be determined by the intensity distributions over all the captured raw images, which can be expressed as follows:

$$I_{th} \leq \overline{\sum_i I_{\max,i}} - \overline{\sum_i I_{std,i}} \qquad (5)$$

Where $I_{\max,i}$ and $I_{std,i}$ denote the maximum and standard deviation of the intensities in the $i^{th}$ captured raw image. In practice, a lower threshold is preferred to reserve more meaningful signals. However, the threshold selection depends on different levels of residual noise, which should be considered according to the specific conditions.

## 3. EXPERIMENT RESULTS AND DISCUSSIONS

The basic experimental setup of our established FPM system is as indicated in Fig. 3. The inverted system is equipped with a 4×, 0.1NA apochromatic objective lens and an 8-bit CCD camera with a pixel size of 3.75μm (DMK23G445, Imaging Source Inc., Germany). A 32×32 programmable RGB LED matrix with 4mm spacing is placed at 86mm above the sample stage. In our experiments, the center 15×15 red LEDs, with an excitation wavelength of 631.13nm and 20nm bandwidth, are carried out to provide angle-varied illuminations, resulting in a synthetic NA of 0.5 theoretically.

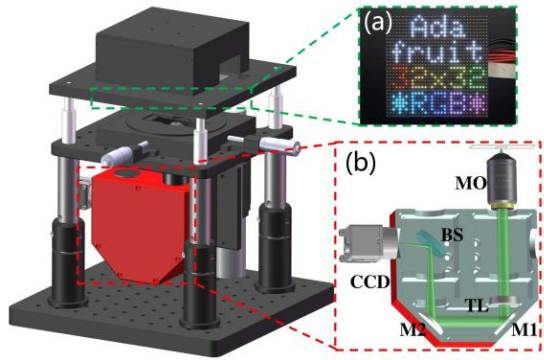

**Fig. 3** Schematic of experimental setup. (a) 32×32 programmable RGB LED matrix. (b) The enlargement of an inverted microscope. MO: microscope objective, TL: tube lens, M1 and M2: mirrors, BS: beam splitter.

Firstly, the effectiveness of step A and step B are validated by a group of contrast experiments as shown in Fig. 4, which presents the reconstruction results of one segment (90×90 pixels) in a USAF target under different treatments. Fig. 4(a) indicates the field of view (FOV) of the target captured with a 4×, 0.1NA objective lens, and (a1) shows a zoom-in image of a sub-region of (a). Groups (b), (c) and (d) denote the reconstructed intensity, phase and spectrum respectively by EPRY-FPM algorithm at 30 iterations, which is the most favored and widely used algorithm for their fast convergence, high computational efficiency and low memory cost. As comparisons, the dataset for image reconstruction is unprocessed in (b1)~(d1), while processed with direct background subtraction in (b2)~(d2) as Fig. 2(c), uniformity method in (b3)~(d3), and both uniformity method and stray light elimination in (b4)~(d4). The reconstructions that without any preprocessing procedures are greatly degraded by the intense noise and stray light with a poor signal-to-noise ratio and severe spectral artifacts, which is as indicated in Fig. 4 (b1)~(d1). And the background subtraction also contributes little to the improvement of recovery quality as demonstrated in (b2)~(d2). However, it could be found a continuous improvement of the intensity images in Fig. 4 (b3) and (b4) with the stepwise implemented step A and step B, and the spectral artifacts in (d3) and (d4) are less obvious compared with that in (d1) and (d2) as shown by red arrows, validating the effectiveness of uniformity method as well as stray light elimination.

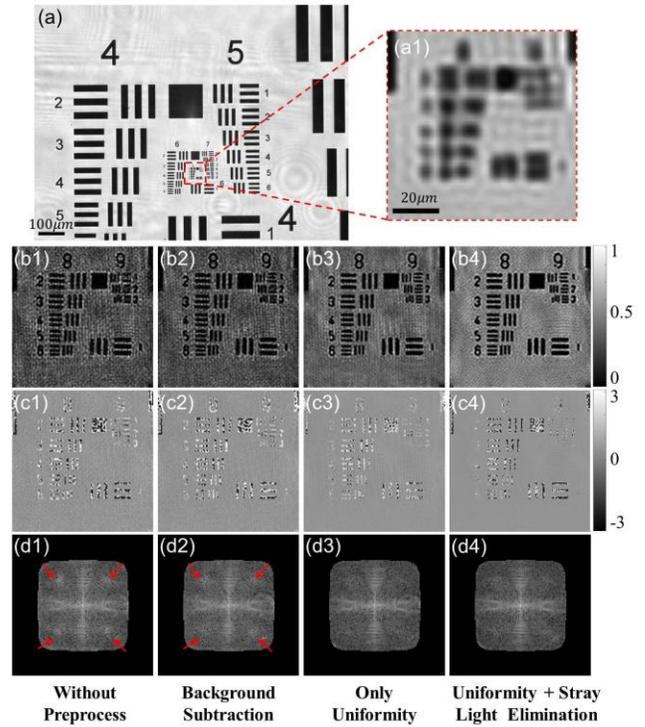

**Fig. 4** Reconstruction results of one segment (90×90 pixels) in a USAF target under different treatments. (a) The FOV captured with a 4×, 0.1NA objective lens, and (a1) presents a zoom-in image of a sub-region of (a). Groups (b), (c) and (d) demonstrate the reconstructed intensity, phase and spectrum respectively. The dataset for image reconstruction is unprocessed in (b1)~(d1), while processed with direct background subtraction in (b2)~(d2), uniformity method in (b3)~(d3), and both uniformity method and stray light elimination in (b4)~(d4).

Then, based on the above experiments, we test a set of threshold values to find out the optimal thresholding for step C. Fig. 5 shows a series of reconstruction results of one segment (90×90 pixels) in a USAF target, with the corresponding thresholds ranging from 0 to 0.03 with an increment of 0.005. Groups (a), (b) and (c) show the reconstructed intensity, phase and spectrum respectively. Comparing these recovery images, it can be observed that a low threshold would leave more residual noise, resulting in a poor signal-to-noise ratio and low contrast in the recovered intensity and phase images, which are indicated in Fig. 5(a1), (a2) and (b1), (b2). Further, the residual noise also leads to sever spectral artifacts as shown by red arrows in Fig. 5(c1) and (c2). With the increase of the threshold value, the spectral artifacts fade away and the intensity images are gradually improved.

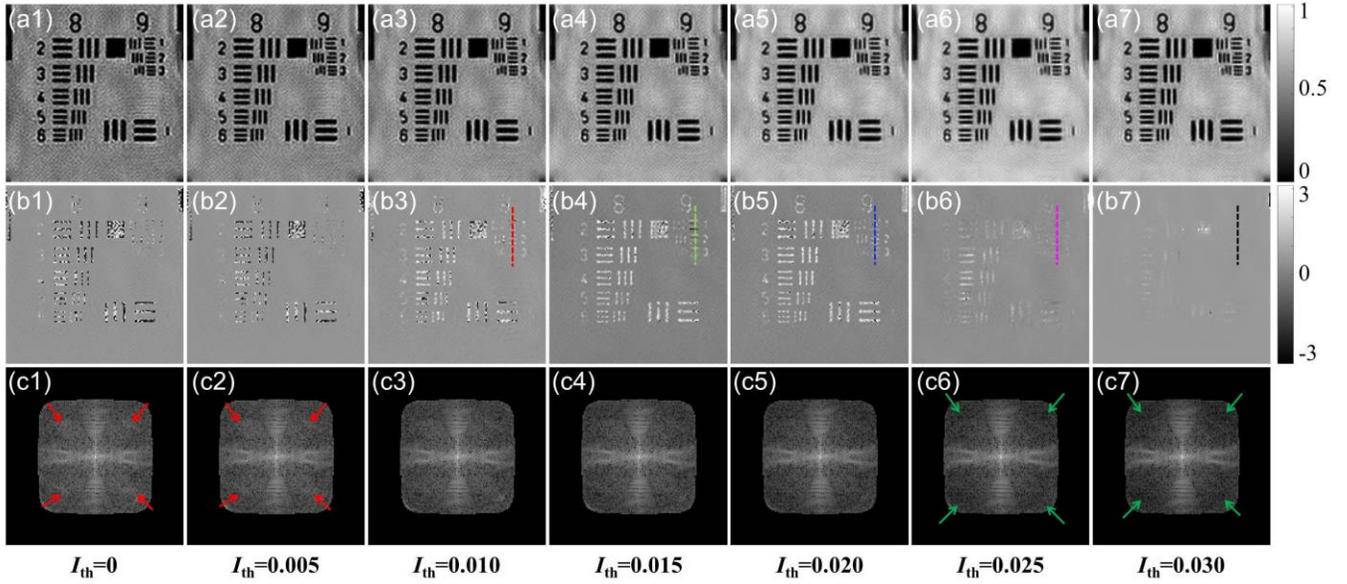

**Fig. 5** Reconstruction results of a USAF target with the corresponding threshold values ranging from 0 to 0.03, with an increment of 0.005. Groups (a), (b) and (c) demonstrate the reconstructed intensity, phase and spectrum respectively by EPRY-FPM algorithm at 30 iterations.

However, a high threshold will erase more meaningful signals, causing a great loss of some high-frequency components. As indicated by green arrows in Fig. 5 (c6) and (c7), the four corners of the spectrum are dimmed, and the corresponding phase images are blurred as demonstrated in (b6) and (b7). Since the phase images present a relatively obvious variation with the increasing thresholds, we measured their intensity distributions along the dotted lines drawn in Fig. 5 (b3)~(b7) to decide the optimal threshold, and the results are presented in Figure 6.

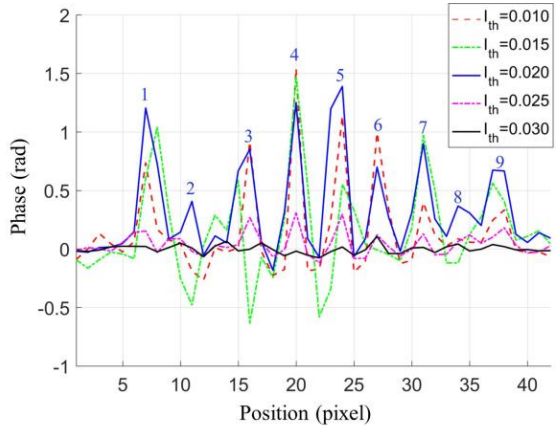

**Fig. 6** The phase distributions with different thresholds along the dotted lines in Fig. 5 (b3)~(b7).

Fig. 6 indicates the phase distributions with different threshold values along the dotted lines in Fig. 5 (b3)~(b7). Here we take the background as the zero-phase point. Comparing these phase curves, the blue curve clearly shows nine sharp peaks along the horizontal axis, with uniform intervals and high drops between the peaks and valleys, which correspond to the nine scribed lines in the USAF target. As for the red and green curves, despite the comparable contrast, they may be subjected to the phase wrapping, which can be seen from that many parts of the two curves are below the zero-phase point. Compared with other phase distributions, the pink and black curves present a poor visibility due to a great loss of high-frequency signals. On the basis of the above analysis, the optimal threshold is supposed to be 0.02 in this experiment.

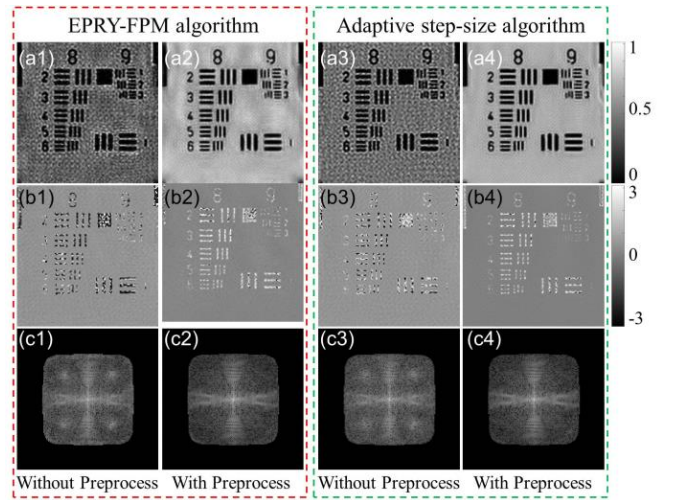

**Fig. 7** Reconstruction results of one segment (90×90 pixels) in a USAF target. Groups (a), (b) and (c) demonstrate the reconstructed intensity, phase and spectrum, and the red and green boxes are recovered by EPRY-FPM and adaptive step-size algorithms respectively, with 30 iterations in (a1)~(c1), 16 iterations in (a3)~(c3), and 18 iterations in the remaining. Here only (a2)~(c2) and (a4)~(c4) are preprocessed.

After the optimal threshold is determined, an overall and systematic data preprocessing scheme is formed. Experimental results of a USAF target that with or without preprocessing, are taken as comparisons to validate the effectiveness of the reported scheme. Fig. 7 groups (a), (b) and (c) demonstrate the reconstructed intensity, phase and spectrum, and the red and green boxes are recovered by EPRY-FPM algorithm and adaptive step-size algorithm [14] respectively, which is one of the recently proposed state-of-the-art algorithms that greatly improves the robustness to noise. Here only (a2)~(c2) and (a4)~(c4) are preprocessed by our reported methods. Through a comparison of these recovery images, we could find that the FP's performance could be markedly enhanced by the provided procedures. It can be observed that Fig. 7(a2) shows a much cleaner background, higher contrast and signal-to-noise ratio than (a1), and the spectral artifacts appeared at four corners in Fig. 7(c1) are also eliminated, demonstrating that the data preprocessing scheme is indeed the most direct and effective way for noise removal. In addition, a better performance could be further obtained by the cooperation with the existing excellent algorithms, which could be seen from that Fig. 7(a4) presents a much more uniform background compared with Fig. 7(a2).

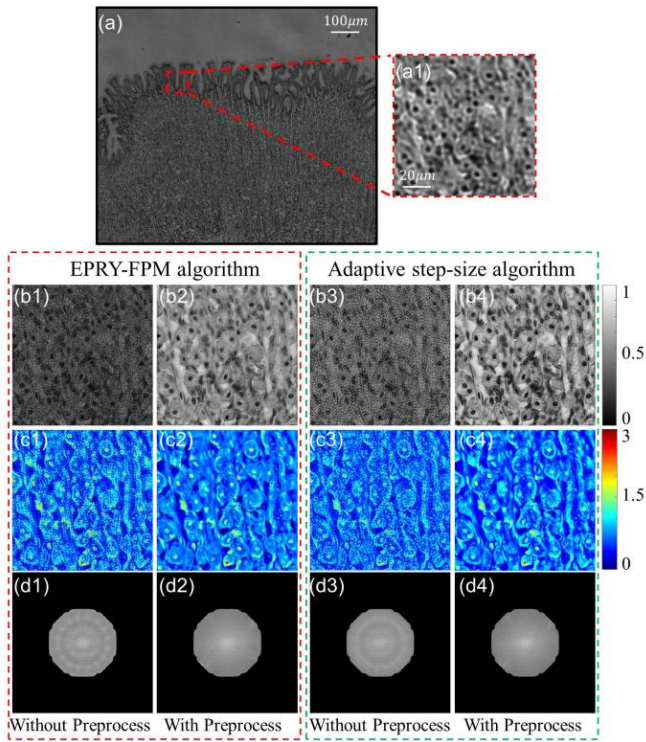

**Fig. 8** Reconstruction results of a dog stomach cardiac region sample. (a) The FOV captured with a 4×, 0.1NA objective lens, and (a1) presents a zoom-in image of a sub-region of (a). Groups (b), (c) and (d) demonstrate the reconstructed intensity, phase and spectrum, and the red and green boxes are recovered by EPRY-FPM algorithm and adaptive step-size algorithm respectively, with 30 iterations in (b1)~(d1), 19 iterations in (b3)~(d3), and 20 iterations in the remaining. Here only (b2)~(d2) and (b4)~(d4) are preprocessed.

In addition, we also test our methods with a public dataset of a biological sample (dog stomach cardiac region), which contains 293 variably-illuminated raw images that captured with a 4×, 0.1NA objective lens and a circular LED matrix. This dataset is available on http://www.laurawaller.com/opensource/, and the corresponding experimental setup is detailed in Ref. [9]. Fig. 8 also compares a set of reconstruction results that with or without preprocessing. Fig. 8(a) presents the field of view (FOV) of the sample and (a1) shows a zoom-in image of a sub-region of (a). Groups (b), (c) and (d) demonstrate the reconstructed intensity, phase and spectrum, and the red and green boxes are recovered by EPRY-FPM algorithm and adaptive step-size algorithm respectively, with 30 iterations in (b1)~(d1), 19 iterations in (b3)~(d3), and 20 iterations in the remaining. Here only (b2)~(d2) and (b4)~(d4) are preprocessed. Through a comparison of these recovery images, the same conclusion can be reached that the reported data preprocessing scheme contributes a lot to the enhancements of FP's performance. Figure 8(b2) shows a much better recovery quality than (b1) and the circular artifacts appeared in (d1) are also eliminated. In addition, the spectral artifacts in Fig. 8(d3) are less obvious than that in (d1), demonstrating the stronger robustness of adaptive step-size algorithm. It can be also found that Fig. 8(b4) further improves the reconstruction quality on the basis of Fig. 8(b2), indicating that the cooperation with the existing state-of-the-art algorithms results in further advancements in final reconstructions.

## 4. CONCLUSIONS

In conclusion, we have proposed an overall and systematic data preprocessing scheme for noise removal from FP's raw dataset, which involves sampling analysis as well as underexposed/overexposed treatments, then followed by the elimination of unknown stray light and suppression of inevitable background noise, especially Gaussian noise and CCD dark current. The reported scheme works successfully on most Fourier ptychographic imaging platforms, without any parameter settings and priori knowledge about noise statistics, facilitating a significant improvement of the FP's performance. Further advancements of reconstruction quality could be obtained by flexibly cooperating with different kinds of state-of-the-art algorithms, producing a stronger robustness of the FP approach in various applications. Experimental results demonstrate that the benefits of noise removal by these methods far outweigh its defects of accompanying signal loss, as part of the lost signals can be compensated by the improved consistencies between the captured images in the process of iterative reconstruction.

**Funding.** Natural Science Fundation of China (NSFC) (61377008, 81427802).

**Acknowledgments**. The authors thank Jiasong Sun for the valuable and helpful discussions and comments.